%% file: root.tex
\documentclass[letterpaper, 10 pt, conference]{ieeeconf}  %

\IEEEoverridecommandlockouts                              %

\usepackage{graphics} %
\usepackage{epsfig} %
\usepackage{mathptmx} %
\usepackage{times} %
\usepackage{amsmath} %
\usepackage{amssymb}  %
\usepackage{caption}

\usepackage{color,soul}
\usepackage{graphicx}

\usepackage{multirow}
\usepackage{array}
\usepackage{booktabs}
\usepackage{makecell} 
\usepackage[table]{xcolor}
\usepackage{highlight}
\usepackage{xspace}
\definecolor{iccvblue}{rgb}{0.21,0.49,0.74}
\usepackage{wrapfig} 
\usepackage{marvosym}
\usepackage[pagebackref,breaklinks,colorlinks,allcolors=iccvblue]{hyperref}

\usepackage{cleveref}
\newcommand{\mysubsub}[1]{\noindent\textbf{#1.}}

\newcommand{\MODELNAME}{VistaBot}

\title{\LARGE \bf
VistaBot: View-Robust Robot Manipulation via \\Spatiotemporal-Aware View Synthesis
}

\author{
Songen Gu$^{1,2*}$,
Yuhang Zheng$^{2,4*}$,
Weize Li$^{2}$,
Yupeng Zheng$^{2,3}$,
Yating Feng$^{2}$,\\
Xiang Li$^{2}$,
Yilun Chen$^{2}$,
Pengfei Li$^{2}$,
Wenchao Ding$^{1,2}\textsuperscript{\Letter}$ \\
\textsuperscript{1}Fudan University,
\textsuperscript{2}TARS Robotics,
\textsuperscript{3}UCAS,
\textsuperscript{4}NUS \\
\textsuperscript{\Letter} Corresponding Author,
\textsuperscript{*} Equal Contribution \\
}

\begin{document}

\maketitle

\thispagestyle{empty}
\pagestyle{empty}

\input{sec/0_abstract}

\input{sec/1_intro}

\input{sec/2_related_work}
\input{sec/3_method}

\input{sec/4_experiment}
\input{sec/5_conclusion}

\bibliographystyle{IEEEtran}
\bibliography{main}

\end{document}

%% file: sec/0_abstract.tex
\begin{abstract}
Recently, end-to-end robotic manipulation models have gained significant attention for their generalizability and scalability. 
However, they often suffer from limited robustness to camera viewpoint changes when training with a fixed camera. 
In this paper, we propose VistaBot, a novel framework that integrates feed-forward geometric models with video diffusion models to achieve view-robust closed-loop manipulation without requiring camera calibration at test time. 
Our approach consists of three key components: 4D geometry estimation, view synthesis latent extraction, and latent action learning. 
VistaBot is integrated into both action-chunking (ACT) and diffusion-based ($\pi_0$) policies and evaluated across simulation and real-world tasks. 
We further introduce the View Generalization Score (VGS) as a new metric for comprehensive evaluation of cross-view generalization. 
Results show that VistaBot improves VGS by 2.79× and 2.63× over ACT and $\pi_0$, respectively, while also achieving high-quality novel view synthesis. 
Our contributions include a geometry-aware synthesis model, a latent action planner, a new benchmark metric, and extensive validation across diverse environments.
The code and models will be made publicly available.

\end{abstract}

%% file: sec/1_intro.tex
\section{Introduction}
\begin{figure}[t]
\centering
\includegraphics[width= \linewidth]{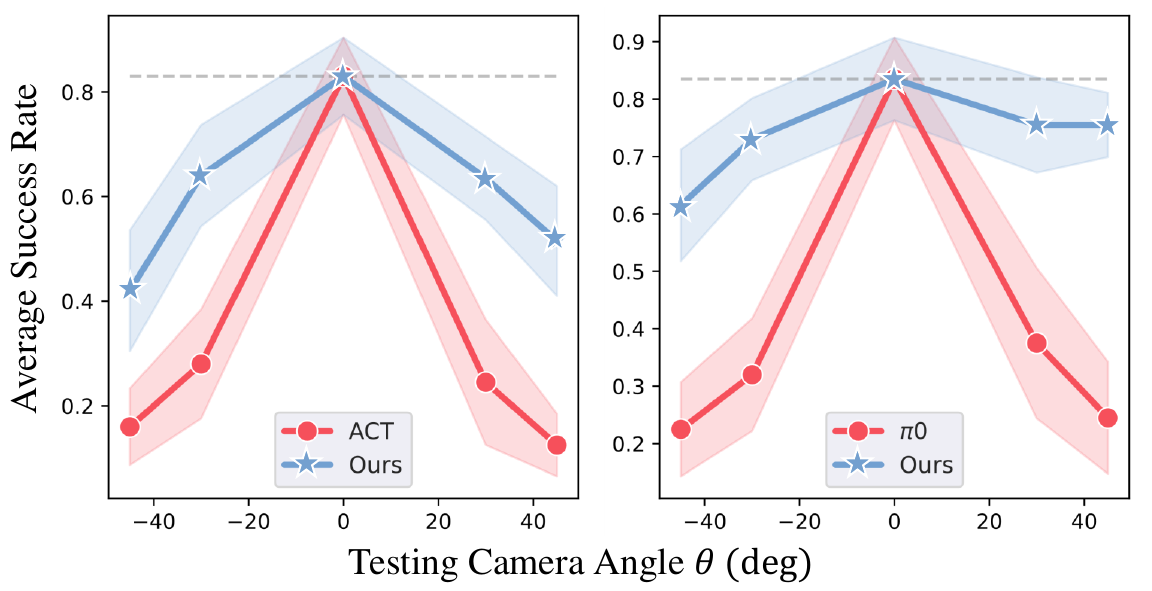}
\caption{\small Our proposed VistaBot demonstrates superior cross-view generalizability compared with SOTA visuomotor policy ($\pi_{0}$ and ACT). As shown in the figure, when the camera observation angle undergoes significant changes, VistaBot consistently maintains a high average success rate even under substantial camera viewpoint changes, whereas the success rates of baseline policies drop to nearly zero as the viewpoint deviates from the training condition.}
    \label{fig:teaser}
    \vspace{-1em}
\end{figure}

End-to-end robotic manipulation have recently gained momentum in robotics, ranging from specialist visuomotor policies trained via imitation learning~\cite{zhao2023learning, chi2023diffusion} to generalist vision-language-action (VLA) models trained on large-scale multi-task datasets~\cite{kim2024openvla, black2024pi_0}. These approaches promise scalability: imitation learning efficiently maps demonstrations to actions, while VLAs aspire to unify instructions and control across diverse tasks.
However, a fundamental bottleneck undermines both paradigms: poor generalization across camera viewpoints. Unlike appearance shifts such as lighting or texture, viewpoint variations disrupt the spatial grounding between perception and action. As a result, even slight changes in perspective can cause dramatic policy failures (Fig.~\ref{fig:teaser}), often forcing practitioners to re-collect demonstrations or retrain models—an outcome directly at odds with the very scalability that end-to-end frameworks claim to deliver.

To address cross-view generalization, prior efforts have primarily fallen into two directions:
(1) Reconstruction-based methods~\cite{zheng2024gaussiangrasper,shorinwa2024splat} attempt to recover the underlying 3D geometry from synchronized multi-view sequences. While conceptually appealing, they are structurally impractical for real-world deployment: precise multi-camera calibration is tedious, and occlusions inevitably lead to unreliable reconstructions.
The tedious process of multi-camera calibration further complicates real-world training and deployment.
(2) Video generation-based methods~\cite{liu2025langscenex,liu2025geometry} leverage multi-view-consistent video generative models and derive actions from predicted future frames. 
However, these typically lack integrated action learning and suffer from low inference efficiency, making them unsuitable for closed-loop robotic manipulation.

In this paper, we present VistaBot, a new framework that fuses feed-forward geometric models with video diffusion models, combining their complementary priors into a unified policy learning pipeline. Unlike prior reconstruction-only or generation-only approaches, VistaBot leverages geometry for structural consistency and diffusion for spatiotemporal completion, yielding a representation that is both physically grounded and visually coherent.
Our key objective is to enable dynamic, closed-loop manipulation that generalizes across camera views—using only a single training viewpoint for demonstrations and arbitrary test viewpoints at inference, without requiring camera calibration or pose information.

Concretely, our framework consists of three key components:
(1) \textbf{4D Geometry Estimation}. We fine-tune a feed-forward geometric model to predict depth and relative camera pose from arbitrary inference frames, aligning test views with the training viewpoint. This enables us to lift 2D observations into 3D point clouds, providing a structural scaffold for cross-view reasoning.
(2) \textbf{Synthesis Latent Extraction}. We employ a conditional video diffusion model (VDM) to consume the reprojected canonical views and encode them into spatiotemporally rich latent features. Furthermore, we introduce a memory mechanism that injects historical latents, yielding a 4D representation that fuses geometry, appearance, and temporal context for closed-loop control.
(3) \textbf{Closed-Loop Action Learning}. We train the policy to operate directly on diffusion latents rather than decoded images, allowing it to learn future actions from representations already imbued with object-level and geometric understanding. This design reduces inference overhead and strengthens the coupling between perception and control.

To rigorously evaluate cross-view generalization in robotic manipulation, we embed VistaBot into two representative end-to-end frameworks: the imitation-learning-based ACT~\cite{zhao2023learning} and the large-scale VLA model $\pi_0$~\cite{black2024pi_0}. We further introduce a new evaluation metric—View Generalization Score (VGS)—to directly quantify policy robustness under viewpoint variation, filling a gap not addressed by existing benchmarks.
In extensive experiments across diverse tasks and both simulation and real-world environments, VistaBot consistently improves VGS by 2.79x over ACT and 2.63x over $\pi_0$, achieving state-of-the-art performance in viewpoint-robust manipulation. In addition, we assess the fidelity of novel view synthesis (NVS), confirming that VistaBot delivers not only robust closed-loop control but also high-quality cross-view reconstructions. Together, these results demonstrate that VistaBot is both effective and broadly applicable, bridging the gap between novel view synthesis and dynamic robotic manipulation.

Our contributions can be summarized as follows:

\begin{itemize}
\item We adapt feed-forward geometric models and video diffusion models to robotic closed-loop control, yielding 4D spatiotemporally consistent latent representations for novel views without requiring camera pose inputs. This removes the dependency on calibration and enables scalable deployment.
\item We propose a latent planner that operates directly on VDM latent features, allowing the policy to learn actions from representations already infused with object-level and geometric understanding. This design enables efficient closed-loop manipulation with strong cross-view generalization.
\item We introduce the View Generalization Score (VGS), a metric that quantitatively measures policy robustness under viewpoint variation, filling a gap in current evaluation protocols.
\item We extensively validate VistaBot across both specialist (ACT) and generalist ($\pi_0$) end-to-end frameworks, in simulation and real-world settings, showing that it delivers robust, state-of-the-art performance in both novel view synthesis and closed-loop manipulation.
\end{itemize}

%% file: sec/2_related_work.tex
\section{Related work}
\begin{figure*}[htb]
\centering
\includegraphics[width= \linewidth]{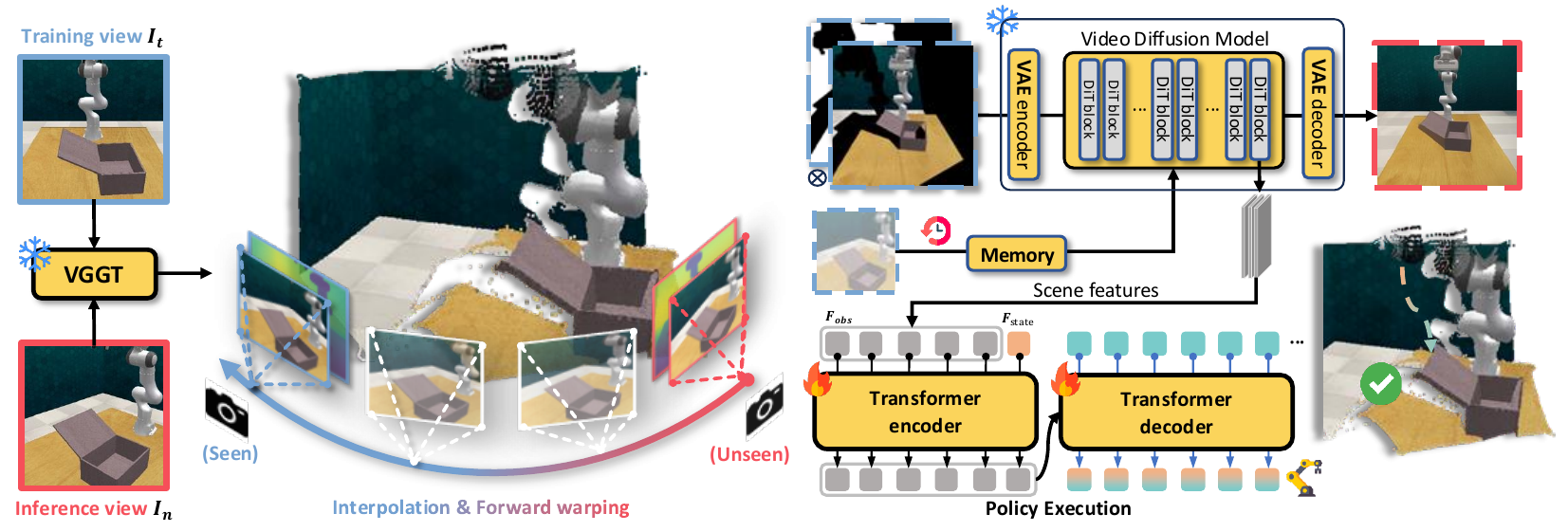}
\caption{\small Architecture of \textbf{VistaBot}. (1) 4D Geometry Estimation with VGGT for pose and depth prediction; (2) View Synthesis via a video diffusion model with memory to generate spatiotemporal-consistent latent features; (3) Policy Execution using a Transformer that fuses scene and robot state features for closed-loop manipulation under unseen views.}
\label{fig:pipeline}
\end{figure*}
\subsection{End-to-End Model for Robotic Manipulation}

With the advancement of visual foundation models~\cite{touvron2023llama, oquab2023dinov2} and large-scale robotic datasets~\cite{o2024open, fang2023rh20t}, end-to-end imitation learning has achieved remarkable progress in robotics. End-to-end policies can be broadly divided into open-loop and closed-loop paradigms. Open-loop approaches~\cite{jaegle2021perceiver,ke20243d} predict keyframes or short-horizon actions without feedback, which makes them scalable across tasks but prone to compounding errors.
Closed-loop visuomotor policies have become the dominant trend. Specialist models such as ACT~\cite{zhao2023learning}, Diffusion Policy~\cite{chi2023diffusion}, DP3~\cite{ze20243d}, and 3D-Diffuser Actor~\cite{ke20243d} achieve strong task-specific performance but require retraining for new skills. In contrast, generalist Vision-Language-Action (VLA) models like OpenVLA~\cite{kim2024openvla} , $\pi_0$~\cite{black2024pi_0}, and recent extensions~\cite{wen2025dexvla, liu2025hybridvla,qu2025spatialvla,cen2025worldvla} leverage large-scale pretraining to support multi-task and zero-shot control.
This progression from open-loop to closed-loop, and from specialists to generalists, marks significant progress. However, both paradigms remain highly sensitive to viewpoint changes, which we address by building viewpoint-robust closed-loop control.

\subsection{Generative Models for Robot Planning}
With the recent advancements in generation models in terms of image quality, temporal consistency, and scene generalization, many works have begun to explore their potential in autonomous driving~\cite{russell2025gaia,gao2023magicdrive,wang2024driving} and robotics~\cite{agarwal2025cosmos}. 
In the realm of robot planning, some works directly employ generative models to produce action videos and then generate the corresponding actions based on the video outputs. These videos can serve as synthesized sub-goals to provide visual guidance for subsequent policy generation~\cite{bu2024closed,cen2024using}, or be utilized to extract robot actions by leveraging inverse dynamics models~\cite{du2023learning, ajay2023compositional, tian2024predictive} or pose tracking models~\cite{liang2024dreamitate, liu2025geometry}. VISTA~\cite{tian2024view}, on the other hand, leverages generative models to synthesize images from novel viewpoints for data augmentation, thereby enhancing the policy's robustness to variations in camera pose.
Another line of research proposes unified models that simultaneously predict both future frames and robot actions~\cite{zhu2025unified,li2025unified,cen2025worldvla}, handling both scene understanding and action generation within a single framework. 
Some approaches leverage generative models as world simulators to support model-based reinforcement learning algorithms~\cite{wu2024ivideogpt,chandra2025diwa} or to enable closed-loop verification of policies~\cite{liao2025genie}.
Rooted in end-to-end robotic manipulation frameworks, our approach aims to produce latent representations from varying observational viewpoints and learn action policies, thereby achieving generalizable manipulation across viewpoints.

\subsection{View-Robust Perception for Robotics}

A long-standing challenge in robotic manipulation is the lack of robustness to camera viewpoint changes.  methods \cite{kerbl20233d,mildenhall2021nerf,zheng2024gaussiangrasper,lu2024manigaussian,liu2025langscenex,seo2024genwarp} , which synthesize unseen views from limited observations,  provide robust perception for robot manipulation. Existing NVS-based approaches generally fall into two categories: reconstruction-based and generation-based. 
Reconstruction-based methods explicitly recover 3D structure from synchronized multi-view data. NeRF- and 3DGS-based methods~\cite{zheng2024gaussiangrasper,lu2024manigaussian} represent the scene as a volume or a 3D Gaussian field and can synthesize high-fidelity novel views. While effective as simulators for offline robot learning, these methods are impractical for closed-loop robotic manipulation
since real-time scene updates of this 3D representation are time-consuming.
Generation-based methods ~\cite{liu2025langscenex,seo2024genwarp} leverage image or video generative models to synthesize novel views for downstream policies. Although they can generate plausible observations, they often struggle with precise camera control since the generation process is not physics-aware.
Unlike the methods above, our approach leverages feed-forward reconstruction to obtain geometric priors and a video generation model to produce viewpoint-aligned spatiotemporal representations, which directly support efficient closed-loop policy learning.

%% file: sec/3_method.tex
\section{Method}
\label{sec:method}
In this section, we introduce \MODELNAME, a framework that combines geometric and video diffusion models for robust manipulation without test-time calibration. We first describe the problem setting in \Cref{sec:Problem_definition}. Then, we present the three key components of our approach: 4D geometry estimation, view-synthesis latent extraction, and latent action learning, detailed in \Cref{sec:Geometry_Estimation}, \Cref{sec:Latent_Extraction}, and \Cref{sec:Action_Learning}, respectively.

\subsection{Problem Formulation}\label{sec:Problem_definition}

To evaluate our proposed policy's generalization ability across different viewpoints, we propose the following task setting: the model is trained exclusively on RGB observations $\{I_t\}$ collected from a fixed camera pose $T_t$. During inference, the model is required to perform closed-loop manipulation tasks based on new observations $\{I_{n}^{i}\}$ captured from a set of novel camera viewpoints $T_{n}^{1}, T_{n}^{2}, \ldots, T_{n}^{m}$, which are distinct from the training viewpoint.

\subsection{4D Geometry Estimation}\label{sec:Geometry_Estimation}
As illustrated in Fig. \ref{fig:pipeline}, given the initial observation of the training view  $I_t$ and an observation of the novel view $I_n$, where $I_t, I_n \in \mathbb{R}^{h \times w \times 3}$, we first utilize a feed-forward geometric model VGGT~\cite{wang2025vggt} to estimate the relative pose $\mathbf{T_{n\to t}}$ between $I_n$ and $I_t$, along with the depth map $D_n$ of $I_n$.
It is noteworthy that since the observation $I_n$ may represent any arbitrary frame during closed-loop operation, directly applying a feed-forward geometric model cannot guarantee temporal consistency. Moreover, existing feed-forward geometric models often underperform on category-specific objects in embodied scenarios, such as the robot arm and gripper. 
To address these limitations, we collect a modest amount of simulated and real-world data to fine-tune the model, enabling consistent 4D geometric estimation in embodied settings.

Next, we lift $I_n$ to obtain its point cloud $P_n \in \mathbb{R}^{hw \times 6}$ with the estimated depth $D_n$ and the camera intrinsic matrix $\mathbf{K} \in \mathbb{R}^{3 \times 3}$.
Then the point cloud is transformed to the training view using the relative pose $\mathbf{T}$ estimated by the feed-forward geometric model, yielding $P_t$:
\begin{equation}
P_t = T_{n\to t}\cdot{\rm proj^{-1}}(I_n, D_n, \mathbf{K}),
\end{equation}
where $\rm proj^{-1}(\cdot,\cdot,\cdot)$ denotes the inverse perspective projection.

Finally, we render $P_t$ into the training view, yielding an image $I_r$:
\begin{equation}
I_r = {\rm proj}(P_t, \mathbf{K}),
\end{equation}
where $\rm proj(\cdot,\cdot)$ represents the perspective projection.

\subsection{Spatial-temporal Latent Feature Extraction}\label{sec:Latent_Extraction}

In this section, our goal is to synthesize the source training view from the point cloud rendering image $I_r$ using a video diffusion model.  
To generate higher-quality source view images and better adapt the video diffusion model for closed-loop robotic control, we design a spatiotemporal conditional strategy, as illustrated in \Cref{fig:pipeline}.  
Finally, we extract the corresponding spatiotemporal latent features from the video diffusion model as scene representations for the policy.

\mysubsub{Point Rendering Inpainting} 
Due to scene occlusions and viewpoint differences, the point cloud rendering $I_r$ contains noticeable holes. We use the projection mask  $M_r$ during the point cloud projection to mark the missing regions. We then use $I_r$ and $M_r$ as conditions to guide a conditional video diffusion model in generating high-quality images at the target training viewpoint. We employ a video diffusion model to inpaint the masked regions. 

\mysubsub{Spatial Viewpoint Interpolation} Image-based inpainting models can fill masked regions with an image and a mask. In our case, however, they attempt to fill large holes caused by substantial viewpoint changes. This process often introduces artifacts and blurred details in the inpainted regions.

To help the generation model capture spatial viewpoint changes and better exploit the contextual information of the video diffusion model, we interpolate camera poses and perform frame-by-frame point cloud rendering to obtain multi-frame interpolated images between the novel and original viewpoints, thereby ensuring a smooth transition during novel-view generation:
\begin{equation}
    \mathbf{T} = t \, \mathbf{T_n} + (1 - t) \, \mathbf{T_t} .
\end{equation}

where $\mathbf{T_n}$ denotes the camera pose of the novel view, and $\mathbf{T_t}$ denotes the source view of the generation target, $t$ is interpolation parameter.

Notably, we adopt CogVideoX \cite{yang2024cogvideox} as the backbone video diffusion model. It consists of a 3D VAE for video latent compression and a DiT for diffusion denoising.  

Specifically, the interpolated images $I_r$ and corresponding masks $M$ are encoded via the VAE and flattened to produce spatial tokens that preserve viewpoint geometry.  

\mysubsub{Temporal Memory Reference}
To maintain temporal consistency of policy observations during closed-loop inference, we introduce a memory module as the temporal condition, where historical references from previous steps are incorporated into subsequent generations.
We cache the observation view frame at the last inference and encode it using the same 3D VAE encoder above. 

Inspired by \cite{yu2025trajectorycrafter}, we adopt a DiT block with cross-attention: spatial tokens serve as queries, while temporal tokens act as keys and values. This design integrates temporal information into the spatial context, yielding 4D-consistent conditions suitable for closed-loop operation.

\begin{figure}[t]
\centering
\includegraphics[width= \linewidth]{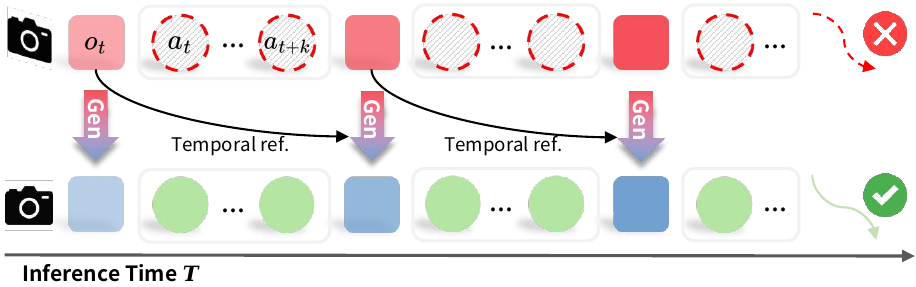}
\caption{\small Closed-loop manipulation during inference. 
Top: unseen-view observations cause action drift and task failure. Bottom: VistaBot combines unseen-view observations with historical references to generate the training (seen) view, enabling consistent action prediction and successful task execution.
``Gen'' refers to our view synthesis process. ``$o$'' and ``$a$'' refer to observation and action, respectively.
}\vspace{-1em}
\label{fig:close_loop_infer}
\end{figure}

\subsection{Closed-loop Action Learning}\label{sec:Action_Learning}
Instead of directly using the generated source view as the policy input,  we use the synthesis latent as our policy's input. 
Inspired by Lexicon3D~\cite{man2024lexicon3d}, video diffusion models (VDM) have demonstrated remarkable performance in object-level and geometric understanding tasks, which is crucial for learning cross-view consistent representations. 
Moreover, compared to explicitly using decoded synthesized images, learning action policies directly in the latent space of the VDM is more efficient, since it omits the decoding of the VAE and the vision encoding of policy.

Specifically, we extract scene features from the final DiT block of the diffusion model after adaptive layer normalization. This representation captures high-level spatiotemporal semantics while preserving generation-relevant structure, making it well suited for visuomotor policy learning.

Following ACT~\cite{zhao2023learning}, we replace the scene features originally extracted by ResNet-50~\cite{He2015DeepRL} with the scene features from the diffusion model $\mathbf{F}_{\text{obs}}$, while retaining the original robot state features $\mathbf{F}_{\text{state}}$ and action chunks.

%% file: sec/4_experiment.tex
\section{Experiments}

In this section, we first describe our experimental setup and introduce the generalization score metric, which is used to evaluate the policy’s ability to generalize across view variations. We then assess the effectiveness of our method through both simulation and real-world experiments. Finally, we conduct an ablation study to validate the key design, followed by a discussion. Collectively, these experiments demonstrate the effectiveness of our method in enhancing view generalization in robotic manipulation policies.

\begin{figure}[htb]
    \centering
    \includegraphics[width= \linewidth]{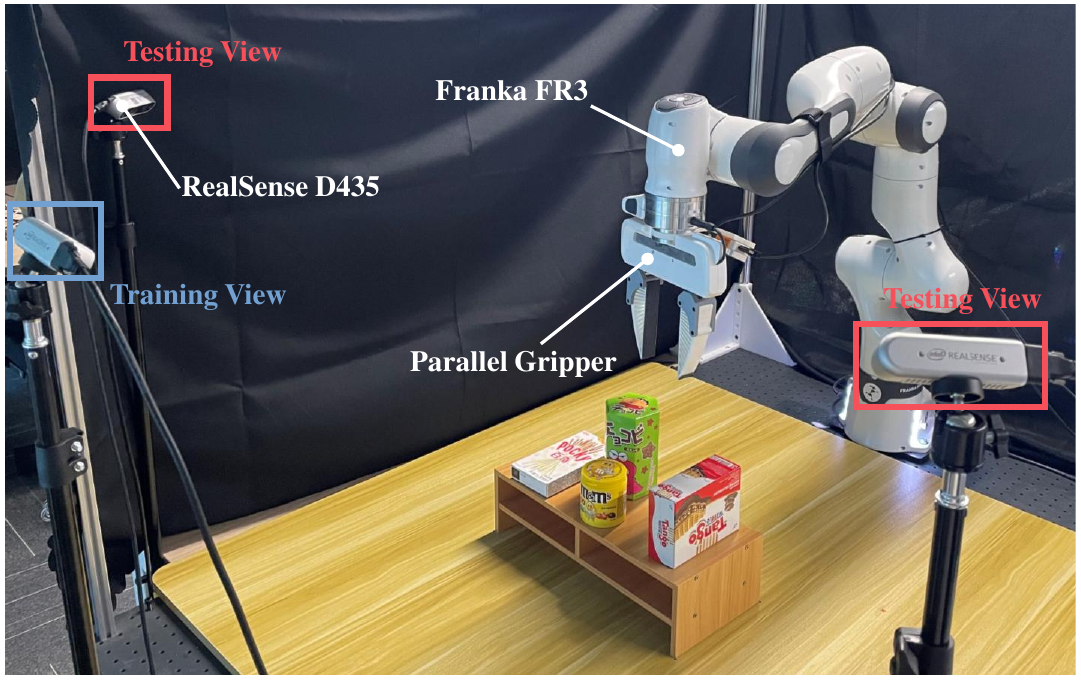}
    \caption{\small Real-world setup with a Franka FR3 arm and three RealSense D435 cameras. The front camera provides training observations, while the side cameras serve as unseen testing views to evaluate cross-view generalization.}
    \label{fig:real_robot}
    \vspace{-1em}
\end{figure}

\input{tab/rlbench_success_rate}

\begin{figure*}
\centering
\includegraphics[width=\linewidth]{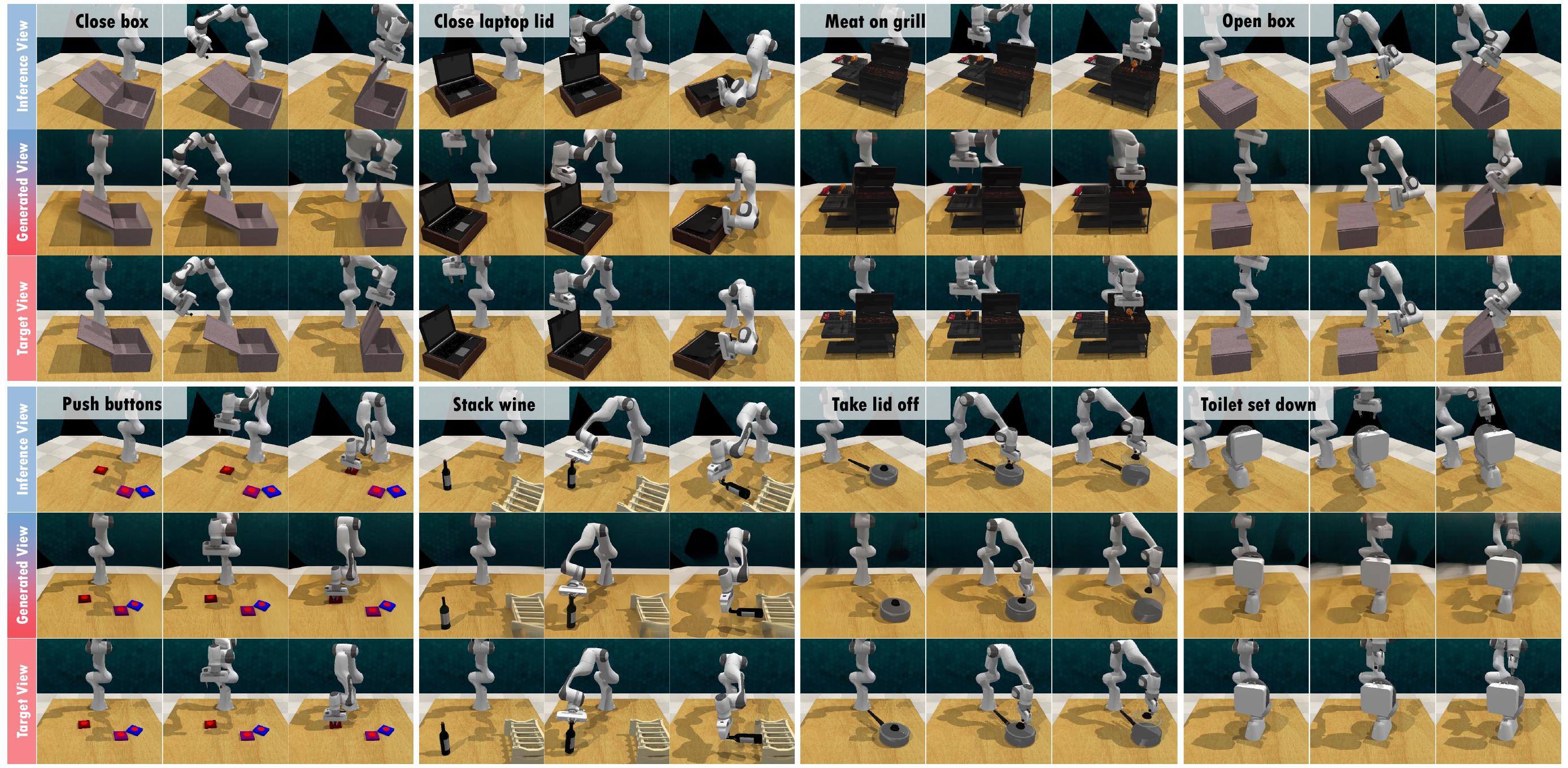}
\caption{\small \textbf{Qualitative results on RLBench.} For each task, VistaBot synthesizes the training viewpoint (middle row) from a novel inference perspective (top row), which is then compared against the ground-truth training viewpoint (bottom row).}
\label{fig:gen_quality_fig}
\end{figure*}

\subsection{Experimental Setup}
To ensure the reproducibility and benchmarking of our experiments, we first conducted the experiments in RLBench~\cite{james2020rlbench} simulated tasks. Subsequently, we also set up a series of tabletop robotic manipulation tasks to evaluate the robustness of our method to view variations in real-world environments. In addition, we proposed a viewpoint generalization metric, \textit{``View Generalization Score"}, to more comprehensively evaluate the robustness of manipulation policies to viewpoint shifts during testing.

\mysubsub{Simulation} We select 8 tasks that are chosen to cover diverse and representative manipulation scenarios from the RLBench tasksuites~\cite{james2020rlbench}, as shown in Fig.~\ref{fig:gen_quality_fig}. Unlike previous approaches~\cite{shridhar2023perceiver, ze2023gnfactor} that regress the discrete keyframe action and manipulate in an open-loop manner, our method employ a close-loop visuomotor policy. During the training stage, we collected 50 demonstrations for each task, where the visual inputs consisted exclusively of RGB images from the front camera. In the testing phase, we used RGB observations from camera viewpoints at $-45^\circ$, $-30^\circ$, $30^\circ$, and $45^\circ$ as inputs, where the rotation angle is defined around the upright axis centered in the robot’s manipulation space. For each testing viewpoint, we evaluated our policy 25 times.

\mysubsub{Real Robot} As shown in Fig.~\ref{fig:real_robot}, we constructed a real-world tabletop robotic platform consisting of a 7-DoF Franka arm equipped with a parallel gripper and three D435 cameras. We selected 4 tasks to evaluate the robustness of our method in real-world scenarios. Similar to the simulation experiments, for each task, we collected 50 trajectories with the observation from the front camera for training. During testing, we evaluated our policy 25 times under each of the left and right camera (novel viewpoints). The whole pipeline predicts robot commands around 3 Hz.

\mysubsub{Baselines} To validate the effectiveness of our method in improving cross-view generalization of manipulation policies and generating high-quality novel-view observations, we set up two groups of baseline comparisons: \textbf{(1) Base visuomotor policies:} We use ACT~\cite{zhao2023learning} and $\pi_0$~\cite{black2024pi_0} as representative generalist and specialist manipulation models, respectively, as our base policies. \textbf{(2) Novel-view synthesis baselines:} We select the state-of-the-art novel-view synthesis methods, Anysplat~\cite{jiang2025anysplat} and LangScene-X~\cite{liu2025langscenex}, as baselines to evaluate the effectiveness of our approach in generating novel-view.

\mysubsub{Evaluation Metrics} To quantitatively evaluate the robustness of our method in cross-view manipulation and the quality of novel-view observation generation, we design three groups of metrics:

\textbf{(1) Average Success Rate (Avg. S.R):} The average task success rate across all tasks under each testing viewpoint.

\textbf{(2) View Generalization Score (VGS):}
We introduce a novel metric, View Generalization Score (VGS), to quantify the robustness of policy performance under viewpoint changes. Let $S(\theta)$ denote the success rate under viewpoint $\theta$, and $S(\theta_0)$ the success rate under the baseline (training) viewpoint $\theta_0$. Given $N$ sampled viewpoints ${\theta_i}{i=1}^N$ within the predefined range $\Theta$, VGS is defined as:
\begin{equation}
\mathrm{VGS} = \frac{1}{N}\sum_{i=1}^N \frac{S(\theta_i)}{S(\theta_0)}.
\end{equation}
VGS measures the average relative performance across novel viewpoints compared to the baseline view, with values closer to $1$ indicating stronger robustness to viewpoint variations.

\textbf{(3) Image and video quality metrics:} We adopt FVD, FID, SSIM, PSNR, and LPIPS to comprehensively evaluate the quality of generated novel-view observations.

\subsection{Simulation Experiments}

We first evaluate the impact of novel testing views on the base policies, as shown in Table~\ref{tab:rlbench}. Compared with the unchanged view (0$^\circ$), testing on novel views (default rows) results in a substantial drop in closed-loop success rate, which is roughly proportional to the magnitude of the viewpoint shift. This trend is observed for both ACT and $\pi_0$. For example, the average success rate of ACT decreases from 0.80 to 0.13 under a 45$^\circ$ view change, representing a reduction of approximately 84\%.  

The VGS score provides a measure of policy view generalizability. The baseline ACT achieves a VGS score of only 0.24, indicating weak view generalizability, while $\pi_0$ performs slightly better with a score of 0.33. By applying our method, both VGS scores and average task success rates improve significantly: the VGS score rises from 0.24 to 0.67 for ACT and from 0.33 to 0.87 for $\pi_0$, demonstrating robust view generalizability under camera viewpoint changes.

We further compare the visual quality of our method with other novel-view synthesis approaches. For reconstruction-based methods, we select the feed-forward approach AnySplat, and for generation-based methods, we select LangScene-X. For AnySplat, we use four views ($-45^\circ$, $-30^\circ$, $30^\circ$, and $45^\circ$) as inputs to reconstruct the 3D Gaussian and render it from the $0^\circ$ viewpoint. For LangScene-X, we use $-45^\circ$ and $45^\circ$ as the first and last frames, then perform video generation to produce an interpolated video; the center frame of this video is used for evaluation.  

\begin{figure}[htb]
    \centering
    \includegraphics[width=0.95 \linewidth]{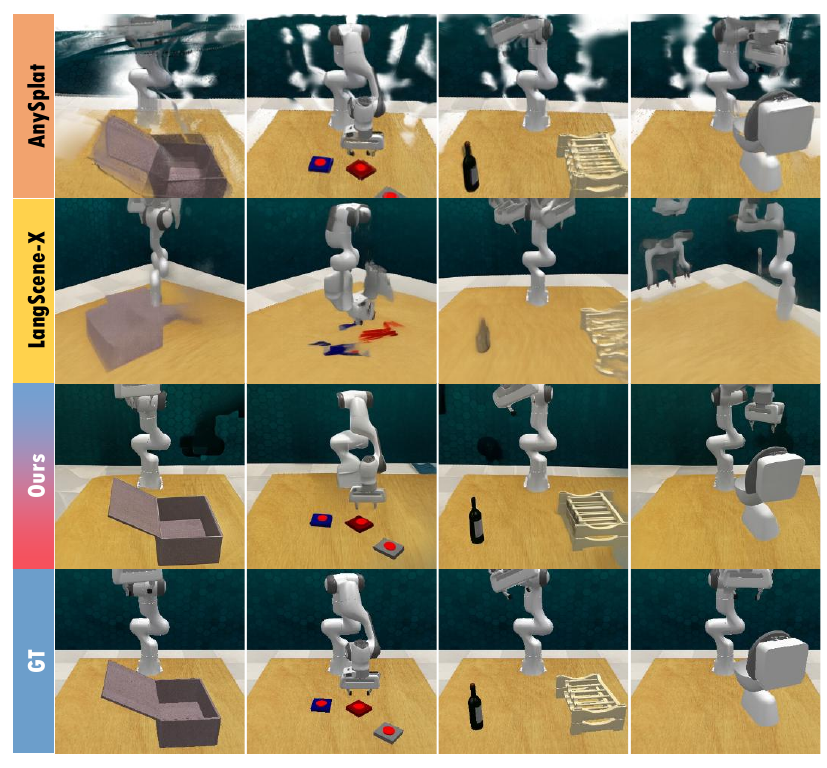}
    \caption{\small Unseen-to-seen view synthesis comparison. \textbf{VistaBot} (Ours) generates sharper and more consistent results than AnySplat and LangScene-X, closely matching the ground truth (GT).}
    \label{fig:cmp_method}
\end{figure}

The generated novel views are shown in Fig.~\ref{fig:cmp_method}. Even when provided with more input views (four in total), AnySplat fails to fill background holes. The limited input also leads to floating artifacts and blurred edges on the robot arm. On the other hand, LangScene-X struggles to accurately estimate the relative poses of the first and last frames, resulting in less precise viewpoints in the target view and often producing “melted” frames that blend the two inputs.  

Our method, by contrast, combines the strengths of both reconstruction and generation. Geometric priors and target camera poses are obtained from VGGT, while unseen regions are filled using the video diffusion model, resulting in plausible visual quality for novel views.

\input{tab/image_quality}

\subsection{Real-World Experiments}
\begin{figure}
\centering
\includegraphics[width=0.8\linewidth]{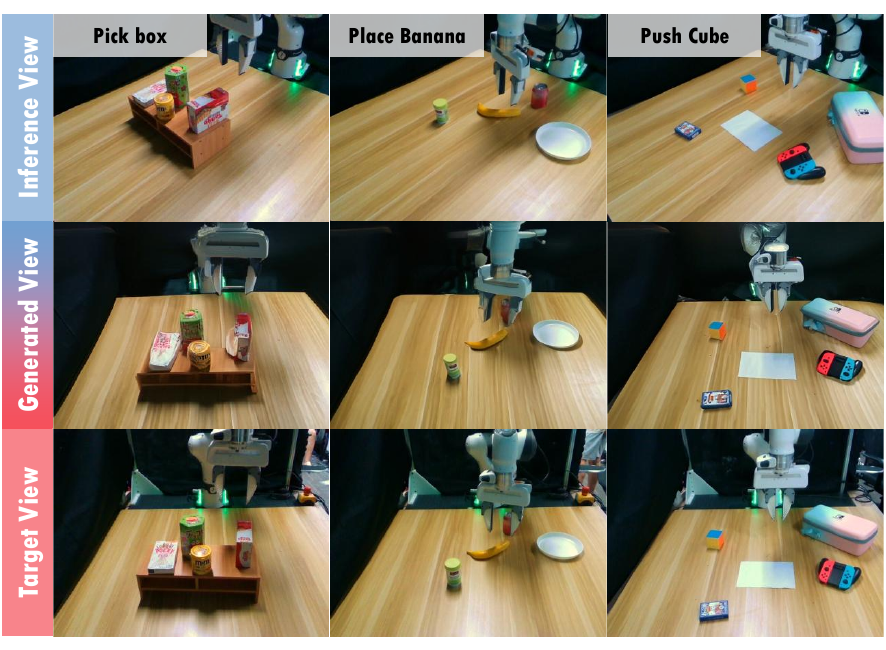}
\caption{\small Qualitative results on Real-robot experiments. }
\label{fig:real}
\vspace{-2em}
\end{figure}

Fig.~\ref{fig:real} presents the results of our real-world experiments, including three distinct tasks. 
The top row displays the viewpoint in test time, the bottom row corresponds to the training viewpoint, and the middle row showcases the training viewpoint generated by our proposed method. These qualitative results demonstrate the strong novel view synthesis capability of our method in real-world environments. Additionally, the quantitative results are shown in the Table~\ref{tab:real_res}. It can be seen that our method exhibits strong cross-view generalization ability in real robot manipulation experiments.

\vspace{-2mm}

\begin{table}[htb]
\centering
\renewcommand{\arraystretch}{1.2}
\setlength{\tabcolsep}{6pt}
\small
\caption{\small \textbf{Quantitative evaluation on real robot.} Average Success rates across 4 tasks under different inference view settings.}
\label{tab:real_res}
\resizebox{\linewidth}{!}{
\begin{tabular}{c|c|c|cccc|c|c}
\toprule
\multirow{2}{*}{\makecell{Base\\Policy}} & \multicolumn{2}{c|}{\textbf{Inference}} & \multicolumn{4}{c|}{\textbf{Tasks}} & \multicolumn{2}{c}{\textbf{Metrics}} \\
\cline{2-9}
 & \textbf{Setting} & \textbf{Angle}
 & \makecell{box on\\shelf} 
 & \makecell{place\\banana} 
 & \makecell{push\\cube} 
 & \makecell{unplug\\charger}
 & \textbf{Avg. S.R.} & \textbf{Avg. VGS} \\
\midrule
\multirow{5}{*}{\textbf{ACT} \cite{zhao2023learning}} 
  & -- & 0$^\circ$ & \g{0.80} & \g{0.88} & \g{0.56} & \g{0.92} & \g{0.79} & --\\
\cline{2-9}
  & \multirow{2}{*}{Default} 
    & -45$^\circ$ & \g{0.12} & \g{0.20} & \g{0.04} & \g{0.24} & \g{0.15} & \multirow{2}{*}{0.21} \\
  &  & +45$^\circ$ & \g{0.24} & \g{0.16} & \g{0.08} & \g{0.24} & \g{0.18} & \\
\cline{2-9}
  & \multirow{2}{*}{Ours} 
     & -45$^\circ$ & \g{0.32} & \g{0.60} & \g{0.48} & \g{0.72} & \g{0.53} & \multirow{2}{*}{\textbf{0.72}} \\
  &  & +45$^\circ$ & \g{0.64} & \g{0.64} & \g{0.48} & \g{0.68} & \g{0.61} &\\
\midrule
\multirow{5}{*}{\textbf{$\pi_0$} \cite{black2024pi_0}} 
  & -- & 0$^\circ$ & \g{0.92} & \g{0.92} & \g{0.64} & \g{1.00} & \g{0.87} & --\\
\cline{2-9}
  & \multirow{2}{*}{Default} 
    & -45$^\circ$ & \g{0.16} & \g{0.28} & \g{0.12} & \g{0.32} & \g{0.22} & \multirow{2}{*}{0.27} \\
  &  & +45$^\circ$ & \g{0.32} & \g{0.28} & \g{0.16} & \g{0.36} & \g{0.28} &\\
\cline{2-9}
  & \multirow{2}{*}{Ours} 
    & -45$^\circ$ & \g{0.40} & \g{0.80} & \g{0.52} & \g{0.88} & \g{0.65} & \multirow{2}{*}{ \textbf{0.79} } \\
  &  & +45$^\circ$ & \g{0.68} & \g{0.76} & \g{0.56} & \g{0.88} & \g{0.72} & \\
\bottomrule
\end{tabular}
}
\vspace{-3mm}
\end{table}

\subsection{More Discussion}
\mysubsub{Ablation Study}
Table~\ref{tab:ablation} presents the results of our ablation study. Compared with Row 3, our approach using estimated 4D geometry closely approximates the performance achieved with ground-truth (GT) geometry. In contrast to Row 4, the results underscore the contribution of the memory module to effective closed-loop control.

\input{tab/ablation}

\begin{figure}[htb]
    \centering
    \includegraphics[width= 0.8\linewidth]{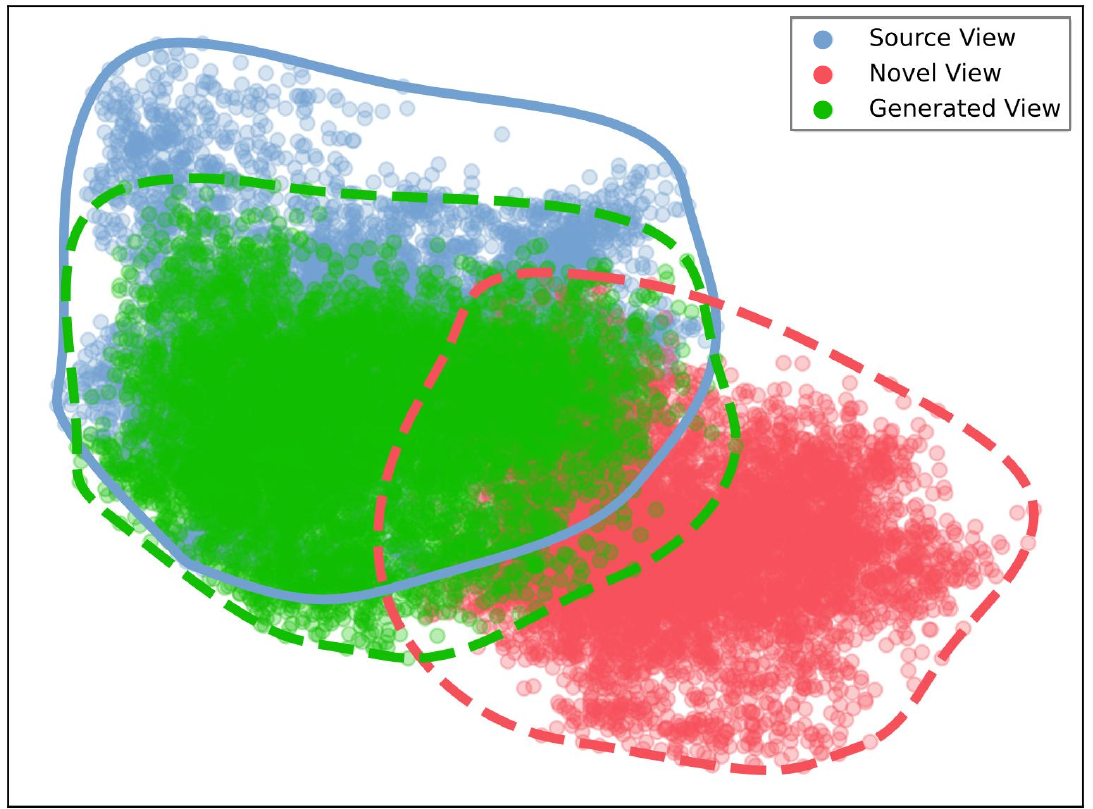}
    \caption{\small Visualization of feature distributions from different viewpoints in the principal component space. Novel views (Red) diverge from the source view (Blue), while \textbf{VistaBot}-generated views (Green) align closely with the source, highlighting improved cross-view generalization.}
    \label{fig:distribution}
    \vspace{-1em}
\end{figure}

\mysubsub{Estimation of View Generalizability}
We now examine in detail what happens when testing under view disturbances, why this degrades policy performance, and how our method alleviates this issue.  
Since policy learning typically relies on a visual encoder (e.g., ResNet in ACT or ViT in $\pi_ 0$) to capture environment observations, we encode the source training view, novel testing view, and the generated view from our method using a pretrained InceptionV3 across the dataset. We then apply PCA to reduce dimensionality, and the resulting features are shown in Fig.~\ref{fig:distribution}.

It is evident that when a novel view is used during testing, the feature distribution shifts significantly from the source training view. This distributional shift induces an out-of-distribution (OOD) problem for the action policy, causing the estimated actions to deviate from the correct task trajectory. Such deviations further exacerbate the OOD issue, ultimately leading to failure in closed-loop inference.  
In contrast, when using our generation model, the distribution of the warped observation features remains much closer to the source training view. As a result, the action head produces outputs that are more consistent with the trained policy, thereby mitigating task failure and preserving task completion capability.

In summary, the key to view generalizability in image-based imitation learning lies in maintaining stable visual representations under view changes. Our proposed method achieves this by producing feature distributions that align more closely with the source training view, thereby enhancing view generalizability.

%% file: tab/rlbench_success_rate.tex
\definecolor{low}{HTML}{B2182B}  %
\definecolor{mid}{HTML}{FEE08B}  %
\definecolor{high}{HTML}{1A9850} %
\newcommand{\g}[1]{\gradientcelld{#1}{0}{0.5}{1}{low}{mid}{high}{50}}

\begin{table*}[htb]
\centering
\renewcommand{\arraystretch}{1.2}
\setlength{\tabcolsep}{6pt}
\caption{\small \textbf{Quantitative evaluation on RLBench.} Success rates across 8 tasks under different inference views.
Green cell color indicates better performance, while red indicates the opposite.
}
\label{tab:rlbench}
\resizebox{0.9\linewidth}{!}{
\begin{tabular}{c|c|c|c c c c c c c c |c |c}
\toprule
\multirow{2}{*}{\makecell{Base\\Policy}} & \multicolumn{2}{c|}{\textbf{Inference}} & \multicolumn{8}{c|}{\textbf{Tasks}} & \multicolumn{2}{c}{\textbf{Metrics}} \\
\cline{2-13}
 & \textbf{Setting} & \textbf{Angle}
 & \makecell{close\\box} 
 & \makecell{close\\laptop} 
 & \makecell{meat on\\grill} 
 & \makecell{open\\box} 
 & \makecell{push\\buttons} 
 & \makecell{stack\\wine} 
 & \makecell{take lid\\off} 
 & \makecell{toilet seat\\down} 
 &  \textbf{Avg. S.R.} & \textbf{Avg. VGS} \\ 
\midrule
\multirow{9}{*}{\textbf{ACT} \cite{zhao2023learning}} 
  & -- & 0$^\circ$ & \g{0.84} & \g{0.60} & \g{0.84} & \g{0.96} & \g{0.64} & \g{0.80} & \g{0.96} & \g{1.00} & \g{0.83} & --\\
\cline{2-13}
  & \multirow{4}{*}{Default} 
    & -45$^\circ$ & \g{0.20} & \g{0.20} & \g{0.00} & \g{0.08} & \g{0.08} & \g{0.00} & \g{0.32} & \g{0.40} & \g{0.16}   &  \multirow{4}{*}{0.24 
}  \\
  &  & -30$^\circ$ & \g{0.32} & \g{0.20} & \g{0.00} & \g{0.32} & \g{0.32} & \g{0.08} & \g{0.40} & \g{0.68} & \g{0.28}   & \\
  &  & +30$^\circ$ & \g{0.56} & \g{0.20} & \g{0.04} & \g{0.08} & \g{0.08} & \g{0.04} & \g{0.32} & \g{0.64} & \g{0.25}  &\\
  &  & +45$^\circ$ & \g{0.28} & \g{0.16} & \g{0.00} & \g{0.08} & \g{0.08} & \g{0.00} & \g{0.08} & \g{0.32} & \g{0.13}   & \\
\cline{2-13}
  & \multirow{4}{*}{Ours} 
     & -45$^\circ$ & \g{0.40} & \g{0.36} & \g{0.40} & \g{0.32} & \g{0.52} & \g{0.12} & \g{0.32} & \g{0.92}  & \g{0.42}  & \multirow{4}{*}{\textbf{0.67}}\\
&  & -30$^\circ$ & \g{0.76} & \g{0.44} & \g{0.40} & \g{0.72} & \g{0.64} & \g{0.52} & \g{0.64} & \g{1.00}   & \g{0.64} &\\
&  & +30$^\circ$ & \g{0.60} & \g{0.44} & \g{0.52} & \g{0.76} & \g{0.68} & \g{0.48} & \g{0.68} & \g{0.92}  & \g{0.64}  &\\
&  & +45$^\circ$ & \g{0.56} & \g{0.40} & \g{0.32} & \g{0.28} & \g{0.64} & \g{0.40} & \g{0.60} & \g{0.92}  & \g{0.52}  &\\
\midrule
\multirow{9}{*}{\textbf{$\pi_0$} \cite{black2024pi_0}} 
  & -- & 0$^\circ$ & \g{0.96} & \g{0.84} & \g{0.84} & \g{0.96} & \g{0.68} & \g{0.56} & \g{0.92} & \g{0.92} & \g{0.84} & --\\
\cline{2-13}
  & \multirow{4}{*}{Default} 
    & -45$^\circ$ & \g{0.24} & \g{0.24} & \g{0.04} & \g{0.28} & \g{0.16} & \g{0.00} & \g{0.32} & \g{0.52} & \g{0.23}  & \multirow{4}{*}{0.33}\\
  &  & -30$^\circ$ & \g{0.20} & \g{0.32} & \g{0.24} & \g{0.24} & \g{0.20} & \g{0.12} & \g{0.56} & \g{0.68} & \g{0.32} &\\
  &  & +30$^\circ$ & \g{0.72} & \g{0.60} & \g{0.20} & \g{0.12} & \g{0.32} & \g{0.00} & \g{0.64} & \g{0.40} & \g{0.38} &\\
  &  & +45$^\circ$ & \g{0.60} & \g{0.40} & \g{0.08} & \g{0.16} & \g{0.04} & \g{0.08} & \g{0.24} & \g{0.36} & \g{0.24} &\\
\cline{2-13}
  & \multirow{4}{*}{Ours} 
    & -45$^\circ$ & \g{0.44} & \g{0.60} & \g{0.56} & \g{0.36} & \g{0.76} & \g{0.48} & \g{0.80} & \g{0.92} & \g{0.62} &  \multirow{4}{*}{\textbf{0.87}}\\
  &  & -30$^\circ$ & \g{0.76} & \g{0.60} & \g{0.68} & \g{0.52} & \g{0.72} & \g{0.76} & \g{0.80} & \g{1.00} & \g{0.73} &\\
  &  & +30$^\circ$ & \g{0.92} & \g{0.56} & \g{0.76} & \g{0.92} & \g{0.80} & \g{0.48} & \g{0.72} & \g{0.96} & \g{0.76} &\\
  &  & +45$^\circ$ & \g{0.76} & \g{0.72} & \g{0.72} & \g{0.88} & \g{0.76} & \g{0.52} & \g{0.84} & \g{0.84} & \g{0.76} & \\
\bottomrule

\end{tabular}
}

\end{table*}

%% file: tab/image_quality.tex
\begin{table}[htb]
\centering
\renewcommand{\arraystretch}{1.2}
\setlength{\tabcolsep}{8pt}
\caption{\small {Quantitative results of view synthesis.}}\label{tab:metrics}
\resizebox{0.95\linewidth}{!}{
\begin{tabular}{c|c c c c c}
\toprule
Method & FVD$\downarrow$ & FID $\downarrow$ & SSIM $\uparrow$ & PSNR $\uparrow$& LPIPS $\downarrow$ \\
\midrule
AnySplat \cite{jiang2025anysplat} & 825.83  & 102.71  & 0.27  & 12.07  & 0.23  \\
LangScene-X \cite{liu2025langscenex} & 551.91  & 118.34  & 0.44  & 15.02  & 0.17  \\
Ours & \textbf{471.04} & \textbf{69.56} & \textbf{0.59} & \textbf{18.34} & \textbf{0.09} \\
\bottomrule
\end{tabular}
}
\vspace{-2em}
\end{table}

%% file: tab/ablation.tex
\begin{wraptable}{l}{0.45\linewidth} %
\vspace{-8pt}
\centering
\renewcommand{\arraystretch}{1.2}
\setlength{\tabcolsep}{8pt}
\caption{\small \textbf{Ablation study.}}
\label{tab:ablation}
\resizebox{0.95\linewidth}{!}{
\begin{tabular}{c|c}
\hline
Method & VGS $\uparrow$ \\
\hline
ACT & 0.24 \\
VistaBot (ours)  & 0.67 \\
w/ GT Depth and extr.   & \textbf{0.79} \\
w/o Memory  & 0.48 \\
\hline
\end{tabular}
}

\vspace{-10pt}
\end{wraptable}

%% file: sec/5_conclusion.tex
\vspace{-0.3em}
\section{Conclusion}

In this paper, we present VistaBot, a novel framework for view-robust robot manipulation that integrates feed-forward geometry and video diffusion models. 
Without requiring camera poses, our method achieves spatiotemporally consistent 4D representation learning and closed-loop action inference in novel views. 
We introduced a view generalization score (VGS) for systematic evaluation and showed significant improvements over strong baselines like ACT and $\pi_0$ in both simulated and real-world environments. 

\mysubsub{Limitation}
Although VistaBot enables the synthesis of novel views for robotic manipulation, it has certain limitations. The model struggles to generate high-quality synthesized views under severe occlusions.